\documentclass[lettersize,journal]{IEEEtran}
\usepackage{amsmath,amsfonts}
\usepackage{algorithmic}
\usepackage{algorithm}
\usepackage{array}
\usepackage[caption=false,font=normalsize,labelfont=sf,textfont=sf]{subfig}
\usepackage{textcomp}
\usepackage{stfloats}
\usepackage{url}
\usepackage{verbatim}
\usepackage{graphicx}
\usepackage{cite}
\usepackage{color, xcolor, colortbl}
\definecolor{BoldDelta}{HTML}{287EB8}
\definecolor{CiteBlue}{HTML}{2471a3}
\definecolor{LightGray}{HTML}{F0F1F2} 

\usepackage{amssymb}
\usepackage{hyperref}
\hypersetup{colorlinks,linkcolor={red},citecolor={black},urlcolor={black}}
\usepackage{booktabs}
\usepackage{multirow}
\usepackage{makecell}
\usepackage{algorithm}
\usepackage{algorithmic}

\definecolor{darkpastelgreen}{rgb}{0.13, 0.9, 0.34}

\newcommand{\bfstart}[1]{\noindent\textbf{#1.}}

\newcommand{\ie}{\textit{i}.\textit{e}.}
\newcommand{\eg}{\textit{e}.\textit{g}.}

\newcommand{\actcond}{$\mathrm{cond}$}

\newcommand{\CommetBlue}[1]{{\color[RGB]{45,28,128} #1}}

\begin{document}
\title{Generalization Beyond Feature Alignment: \\Concept Activation-Guided Contrastive Learning}
\author{Yibing Liu,~Chris Xing Tian,~Haoliang Li,~Shiqi Wang,~\IEEEmembership{Senior Membor,~IEEE,}
}

\markboth{Journal of \LaTeX\ Class Files,~Vol.~14, No.~8, August~2021}%
{Shell \MakeLowercase{\textit{et al.}}: A Sample Article Using IEEEtran.cls for IEEE Journals}


\maketitle

\begin{abstract}
	Learning invariant representations via contrastive learning has seen state-of-the-art performance in domain generalization (DG). Despite such success, in this paper, we find that its core learning strategy -- feature alignment -- could heavily hinder model generalization. 
	Drawing insights in neuron interpretability, we characterize this problem from a neuron activation view. 
	Specifically, by treating feature elements as neuron activation states, we show that conventional alignment methods tend to deteriorate the diversity of learned invariant features, as they indiscriminately minimize all neuron activation differences. 
	This instead ignores rich relations among neurons -- {many of them often identify the same visual concepts despite differing activation patterns}.
	With this finding, we present a simple yet effective approach, \textit{Concept Contrast} (CoCo), which relaxes element-wise feature alignments by contrasting high-level concepts encoded in neurons. Our CoCo performs in a plug-and-play fashion, thus it can be integrated into any contrastive method in DG.
	We evaluate CoCo over four canonical contrastive methods, showing that CoCo promotes the diversity of feature representations and consistently improves model generalization capability. 
	By decoupling this success through neuron coverage analysis, we further find that CoCo potentially invokes more meaningful neurons during training, thereby improving model learning. 
\end{abstract}

\begin{IEEEkeywords}
Domain generalization, neuron activation, contrastive learning, concept contrast.
\end{IEEEkeywords}

\section{Introduction}
\label{sec:intro}
\IEEEPARstart{O}{ver} the past decades machine learning systems have achieved remarkable success~\cite{tech:Imagenet,tech:conv,tech:ResNet,tech:deep_conv}, with the assumption that training and test data are sampled from independent and identical distributions. In real-world applications where data acquisition condition largely varies, however, this assumption rarely holds. The distribution shifts between training and test data thus appear, leading to the degraded performance on well-trained models. To this end, domain generalization (DG)~\cite{Setup:DG} proposes to learn models that generalize well on unseen domains by leveraging a collection of source domains during training.

\begin{figure}
	[t]
	\centering 
	\includegraphics[width=0.99\columnwidth]{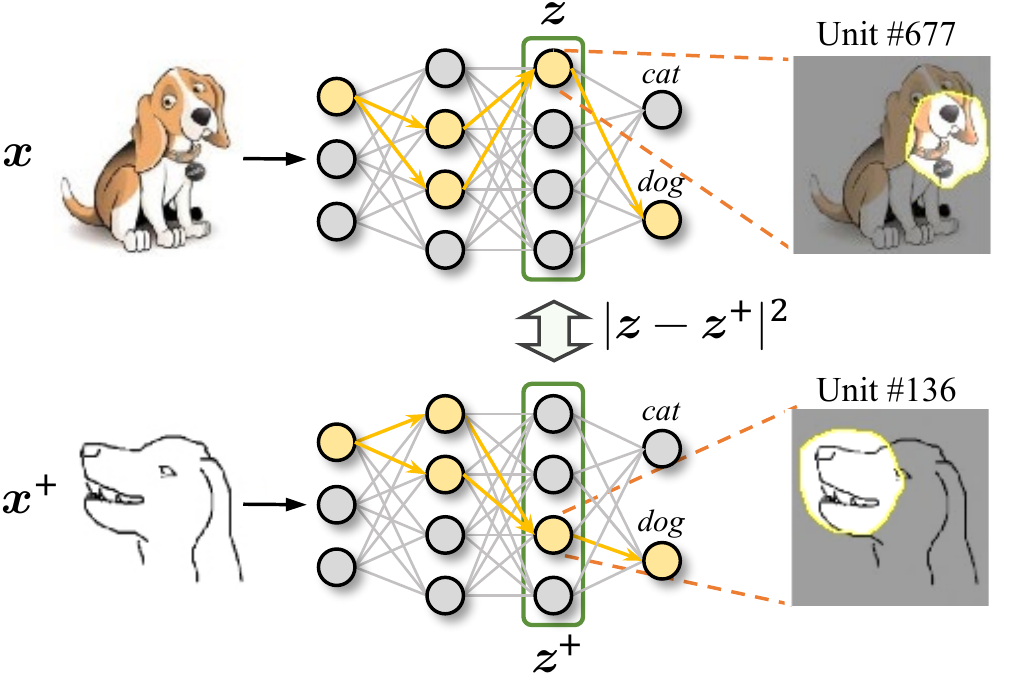} %
	\caption{
		Illustration of contrastive feature alignment from a neuron activation view, where neurons with yellow (gray) color denote activated (non-activated) ones. 
		$\boldsymbol{x}$ and $\boldsymbol{x}^+$ indicate positive samples, and $\boldsymbol{z}$ and $\boldsymbol{z}^+$ are normalized features.
		Since different neurons could identify similar visual concepts (\eg, unit 677 and 136 both detect \textit{dog nose}), we argue that such explicit alignment methods, {which overly stress the same neuron activations} (\ie, $|\boldsymbol{z}-\boldsymbol{z}^+|^2$), could deteriorate the diversity of learned invariant features, further leading to the overfitting problem in the model generalization.
	}
	\label{fig:intro_motivation}
\end{figure}

\begin{figure*}
	[t]
	\centering 
	\includegraphics[width=0.92\textwidth]{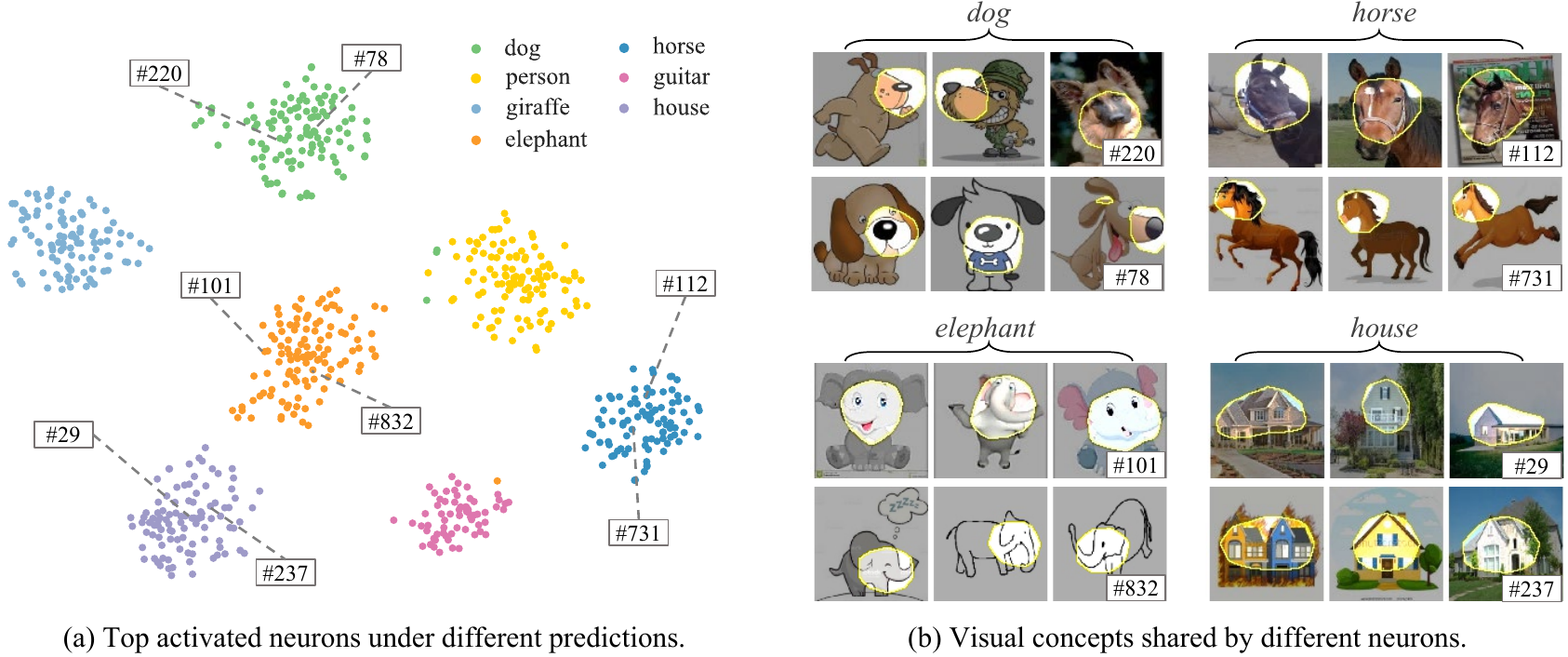} %
	\caption{The connection between neurons and visual concepts. All neurons are obtained from the last convolutional layer in the feature extractor, \ie, ResNet-50. (a) A well-trained ERM model relies on specific groups of neurons to make predictions over the PACS dataset. (b) Among each group, many neurons (\eg, unit 220 and 78) act as detectors and identify the similar visual concepts for model decisions. }
	\label{fig:intro_projection}
\end{figure*}

In contrast to domain adaptation~\cite{Setup:DA} which assumes target data is accessible, DG considers a more realistic scenario where only source domains can be utilized during the training phase.
As such, there is much hope resting on that the learned feature representations could keep invariance on the unseen targets~\cite{Baseline:IRM,Baseline:Fish,Baseline:Fishr,TIP1:DG_LowRank,TIP3:DG_GIN,TIP4:DG_ACE}, thereby powering the predictor trained merely on source domains to perform well. 
Benefiting from the recent advances in self-supervised learning~\cite{CL&Base:SimCL,CL&Base:BYOL,CL:SupCL,TIP7:SSL_Twin}, contrastive-based learning has been proposed as a promising solution~\cite{CL&DG:Unified,CL&DG:MASF(Douqi),CL&DG:CCAI,CL&DG:EISNet,CL&DG:SelfReg,CL&DG:CSG,CL&DG:CasualMatch,CL&DG:PDEA,CL&DG:CondCAD}. 
A canonical paradigm underlying this theme is to exploit rich sample-to-sample relations from different domains, whereby the feature representations of same-class samples (positive pairs) are aligned together, while the different-class ones (negative pairs) are brought apart. 
Since feature distance is explicitly optimized over domain shifts, the model could eventually learn class-conditional feature invariance via such contrasts.



However, in this paper, we find that conventional \textit{feature alignment} strategies of contrastive learning would instead hinder the model generalization. 
To illustrate this point, we employ a neuron activation view inspired by the recent progress in neuron interpretability~\cite{NACT:PNAS_Dissect,NACT:CVPR_Dissect,NACT:ConceptEvo,NACT:Input_Synthesis,NACT:Multifaceted,NACT:Object_Detect,NACT:MILAN,NACT:PAMI_Dissect}. 
As shown in Figure~\ref{fig:intro_motivation}, we treat feature embeddings as neuron activation states, since all feature elements stem from the neuron output~\cite{NACT:PNAS_Dissect,NACT:MILAN}.
Given rich semantic meanings of neurons, however, we find that current alignment strategies perform via \textit{indiscriminately aligning all neuron states}, \ie, $|\boldsymbol{z}-\boldsymbol{z}^+|^2$.
This instead ignores rich relations among neurons: many of them often identify the same visual concepts (See Figure~\ref{fig:intro_projection}) despite being activated in different ways~\cite{NACT:PNAS_Dissect}.
From this standpoint, we argue that such direct alignment methods -- overly stressing the same route in propagations -- could constrain the diversity of neuron activations. This further deteriorates the learned class-conditional invariant features, thus leading to the overfitting problem in model generalization. We provide the empirical analysis of this problem in Figure~\ref{fig:kde_feat_div}.

In this paper, we present \textit{Concept Contrast} (CoCo) to address the above problem. 
Instead of performing element-wise alignments at the feature level, our CoCo contrasts high-level concepts encoded in neurons. 
Concretely, as many neurons in the network often operate in a similar fashion, we first model relations among neurons by employing neuron summarization techniques~\cite{NACT:ConceptEvo,NACT:NeuroCartography}, whereby different neurons identifying the {similar visual concepts} are clustered.
To further retrieve concept representations from features, we then introduce concept activation vectors (CAVs), which aggregates activation states from same-cluster neurons.
Finally, by optimizing the divergence between CAVs, we carry out contrastive loss on the concept level. 
Note that since our method does not impose extra parameters, it can be applied into any contrastive method in DG with easy integration.

To validate the effectiveness of CoCo, we conduct extensive experiments with four typical contrastive methods over various DG benchmarks. The quantitative and qualitative results demonstrate that CoCo not only boosts invariant feature learning, but also consistently improves model performance across datasets and methods. 
In addition, through neuron coverage analysis~\cite{NACT_DG:NeuronCoverage,NACT_System:DeepXplore,NACT_System:DeepGauge}, we further show that CoCo promotes the model to invoke more meaningful neurons in propagations, which justifies the rationality of our method.
To sum up, this work makes the following contributions: 

%
%

(i) We introduce a neuron activation view to understand contrastive learning in DG. We empirically find that element-wise alignment could deteriorate the diversity of learned features, leading to the overfitting problem in model generalization.

(ii) We propose Concept Contrast (CoCo), a simple yet effective approach to mitigate overfitting problem in invariant feature learning. CoCo works in a plug-and-play fashion, thus improving contrastive methods with easy integration.

(iii) We experiment CoCo across contrastive methods, backbones, and datasets, showing that CoCo consistently improves invariant feature learning and model generalization ability. 
By decoupling this success through neuron coverage analysis~\cite{NACT_DG:NeuronCoverage,NACT_System:DeepXplore,NACT_System:DeepGauge}, we further find that CoCo potentially invokes more meaningful neurons during training, e.g., +11.5\% neuron coverage ratio on SelfReg~\cite{CL&DG:SelfReg} over PACS dataset~\cite{Dataset:PACS}. 
Our code will be made publicly available.






\begin{figure*}
	[t]
	\centering 
	\includegraphics[width=0.7\textwidth]{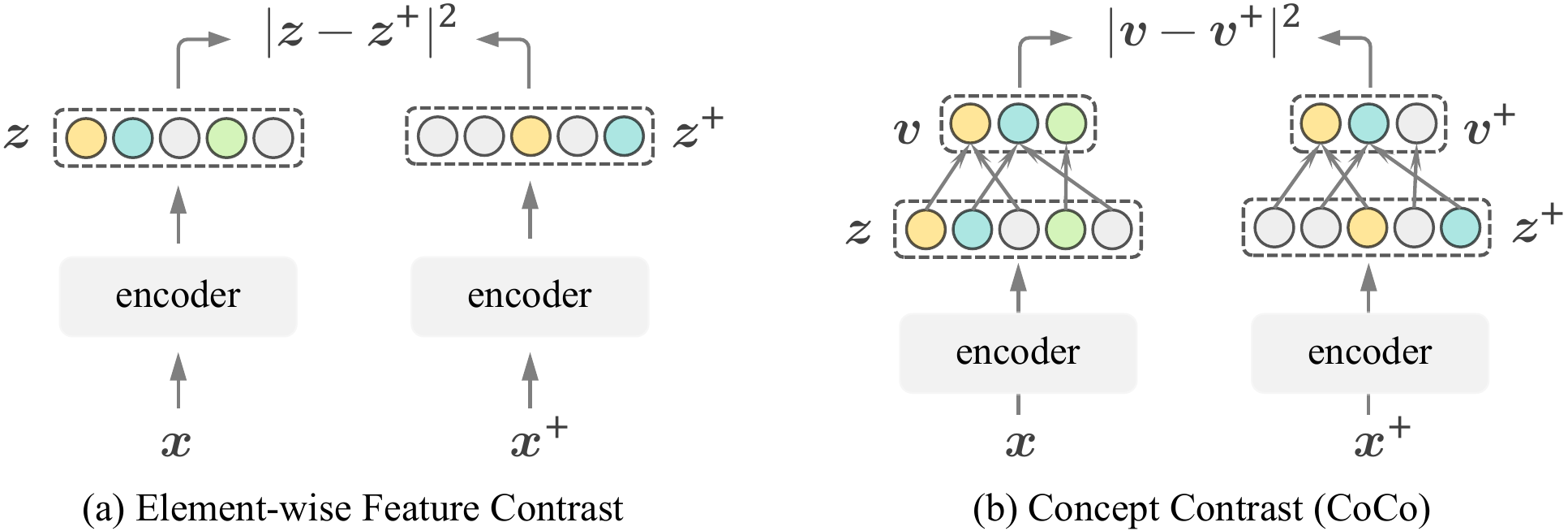} %
	\caption{Conceptual comparison of feature contrast and our concept contrast. Here we utilize three colors (yellow, blue, and green) to differentiate the roles acted by neurons, and gray ones are inactivated neurons.  (a) Element-wise feature contrast requires all neurons to be aligned at the same degree, even though activated neurons (\eg, yellow ones) in two feature vectors have encoded the same visual concepts. (b) CoCo abstracts high-level concepts encoded in neurons and relaxes feature contrast by merely optimizing concept-level differences.}
	\label{fig:method_coco}
\end{figure*}



\section{Related Work}
\bfstart{Domain Generalization} 
The main goal of domain generalization (DG) is to learn models that can overcome distribution shifts between training and test domains. 
Since target data is assumed not accessible under this setting, recent studies mostly center on learning invariant representations over single or multiple source domains~\cite{Baseline:MMD,Baseline:CDANN,Baseline:DANN,Baseline:MixUp,Baseline:CORAL,DG_IRL:Jigsaw,Baseline:IRM,Baseline:V-REx,TIP1:DG_LowRank,TIP3:DG_GIN,TIP4:DG_ACE}. 
In contrast, another  direction attempts to improve generalization via classical learning strategies, such as meta learning~\cite{Baseline:ARM,Baseline:MLDG,TIP2:DG_BAN}, ensemble learning~\cite{DG_Ensemble:GCPR1,DG_Ensemble:ICCV1}, and gradient-based learning~\cite{Baseline:Fish,Baseline:Fishr,NACT_DG:NeuronCoverage}.
Some works also employ data augmentation techniques to enlarge data space and mitigate the overfitting problem in model learning~\cite{DG_Augment:SDG1,DG_Augment:Aug2,DG_Augment:Aug3}.

Specific to contrastive learning in DG, CCSA~\cite{CL&DG:Unified} and MASF~\cite{CL&DG:MASF(Douqi)} are first proposed to improve class-level feature invariance by aligning the feature distributions of same-class samples.
Subsequently, SelfReg~\cite{CL&DG:SelfReg} argues the effects of negative contrast pairs and merely learns features with positive pairs.
PCL~\cite{CL&DG:PCL} further finds that the large distribution gap between source domains could make contrastive learning ineffective, and utilizes proxy-based loss to replace typical simple-to-simple contrast loss. 
On the other end of the spectrum, PDEN~\cite{CL&DG:PDEA} and CSG~\cite{CL&DG:CSG} also apply contrastive learning to single-domain DG for generalized feature learning. 
Contrary to them, in this work, we employ a neuron activation view to reflect contrastive learning in DG and build our method beyond current feature alignment ways.

%


\bfstart{Neuron Interpretability} 
In recent years, there has been much research revealing that a single or combination of neurons detects meaningful information in network propagations~\cite{NACT:CVPR_Dissect,NACT:ConceptEvo,NACT:Summit,NACT:TACV,NACT:Input_Synthesis,NACT:PNAS_Dissect}.
Earlier works such as~\cite{NACT_Base:Patch_Vis1} and~\cite{NACT:Object_Detect} seek to elucidate neuron behaviors by visualizing image patches that maximize neuron activations. 
In line with this, Network Dissection~\cite{NACT:PNAS_Dissect} proposes to systematically compare semantics of individual neurons with visual concepts, where they demonstrated that many neurons in the network often emerge as concept detectors. A recent work in medical analysis also draws a similar conclusion~\cite{NACT:Medical_Neuron}.

Besides understanding single units, another line of research strives to summarize the concepts encoded in collectively activated neurons. For example, TCAV~\cite{NACT:TACV}  vectorizes neuron activations for each layer and measures the layer-wise sensitivity to predefined concepts. 
Summit~\cite{NACT:Summit} and NeuroCartography~\cite{NACT:NeuroCartography} employ interactive view to scalably summarize meaningful neurons and identifies relationship among such neurons. 
Contrary to these work, in this paper, we extend the idea of neuron explainability to feature learning and targets at improving the model generalization ability.


\bfstart{Neuron Coverage} Neuron coverage refers to the percentage of activated neurons given an input set. Motivated by the code coverage in system testing, it is first introduced by~\cite{NACT_System:DeepXplore} to simulate the bug testing in neural networks. The high level idea is that if a network performs with larger neuron coverage during training, there would be fewer undetected bugs (\eg, misclassification) triggered in the test phase. 
In light of this, a number of approaches dive into testing deep learning systems with neuron coverage criteria~\cite{NACT_System:DeepGauge,NACT_System:DeepTest,NACT_System:DeepHunter,NACT_System:PathCoverage}. 
In DG setting, \cite{NACT_DG:NeuronCoverage} recently introduces neuron coverage as a measurement of model generalization capability, and proposes to maximize neuron coverage in the training.
In this paper, we also demonstrate that CoCo has great potential to improve neuron coverage and further improves the model generalization performance.

\section{Methodology}


We consider the common formulation of domain generalization (DG), where multiple source domains are available during training. 
In our formal setup, let $\mathcal{D}=\{{D}_1, {D}_2, ..., {D}_S\}$ be the training set consisting of $S$ source domains on the joint space $\mathcal{X} \times \mathcal{Y}$, where $\mathcal{X}$ and $\mathcal{Y}$ respectively denote the input and label spaces.
We characterize each domain as ${D}_{s} = \{(x_i^s, y_i^s)\}_{i=1}^{N_s}$, where $N_s$ is the number of data samples. 
Without any prior knowledge on the unseen targets, we aim to learn a generalized model $F_{\theta, \phi}$ which comprises of a feature extractor $f_\theta: \mathcal{X} \rightarrow \Gamma$ and a task-related predictor $g_\phi: \Gamma \rightarrow \mathbb{R}^C$, where $\Gamma$ indicates the feature space and $C$ is the number of classes in $\mathcal{Y}$.


In this section, we first motivate our method by reviewing typical contrastive methods in DG, which perform with element-wise feature contrasts (Section \ref{sec_method:contrast_review}). Then, we introduce details of Concept Contrast (CoCo) in Section \ref{sec_method:our_coco}.

\subsection{Preliminaries}
\label{sec_method:contrast_review}
Feature contrast for domain generalization mainly targets at learning class-conditional invariant representations over domain shifts~\cite{CL&DG:Unified,CL&DG:MASF(Douqi),CL&DG:CCAI,CL&DG:SelfReg,CL&DG:CondCAD}.
The key idea is to explicitly optimize the feature distance that aligns the same-class samples from different domains (positive pairs) closer, while pushing different-class ones (negative pairs) apart.
Formally, given an anchor sample $x_i$ within a minibatch, we define its feature embedding as $\boldsymbol{z}_i = f_\theta(x_i)$. 
The positive pair is then constructed by selecting another same-class sample $x_i^+$ in the batch, and negative samples are the remaining ones with different classes. 
By using the typical InfoNCE loss~\cite{CL:InfoNCE}, a feature-level contrast can be given by:
\begin{equation}
	\label{equation:feat_contrast}
	\resizebox{.99\hsize}{!}{
		$\mathcal{L}_f =  -\sum\limits_{i}\mathrm{log}\frac{\mathrm{exp}(-|\boldsymbol{z}_i - \boldsymbol{z}_i^+|^2/2)}{\mathrm{exp}(-|\boldsymbol{z}_i - \boldsymbol{z}_i^+|^2/2)+\sum_{j \ne i} \mathrm{exp}(-|\boldsymbol{z}_i - \boldsymbol{z}_j^-|^2/2)}$
	},
\vspace{1mm}
\end{equation}
where $\boldsymbol{z}_i^+$ and $\boldsymbol{z}_j^-$ are the embedding vectors of positive and negative samples, respectively. 
Note that here $-|\boldsymbol{z}_i - \boldsymbol{z}_i^+|^2/2$ can be directly replaced by the inner product $\boldsymbol{z}_i^T \boldsymbol{z}_i^+$, as all feature embeddings are normalized to unit vectors.

{
	Since all feature elements stem from the neuron output, it is natural to treat feature embeddings as activation strengths of neurons, \ie, activation states~\cite{NACT_DG:NeuronCoverage,NACT:ConceptEvo}. 
	However, given rich relations among neurons (See Figure~\ref{fig:intro_projection}), feature contrast performs in an element-wise way, which aligns all neurons indiscriminately, \ie, $|\boldsymbol{z}_i - \boldsymbol{z}_i^+|^2$ in Equation~\ref{equation:feat_contrast}. 
	This inevitably enforces strong constraints on all neurons, thus limiting representation ability of learned invariant features.}




\subsection{Generalization via Concept Contrast}
\label{sec_method:our_coco}
{
	From the above perspective, we propose concept contrast (CoCo), which relaxes element-wise feature alignments via contrasting high-level concepts encoded in neurons. 
	Our hypothesis is that \textit{as long as two feature vectors encode the same visual concepts, they can be quantified with less difference despite the divergence between their feature elements}. 
	
	We provide a conceptual comparison with feature contrast in Figure~\ref{fig:method_coco}.
	In detail, given an initialized model, CoCo first identifies the relations among neurons, whereby a set of concept clusters $\mathcal{U}^*$ is summarized. 
	Next, to perform the contrastive loss, we abstract features to concept activation vectors (CAV) based on $\mathcal{U}^*$, thereby optimizing divergence at the concept level. 
	We introduce them in the following.
}
\bfstart{Neuron Summarization} The details of neuron summarization are provided in Algorithm~\ref{alg:neuron_summarization}. 
The central idea is to identify concepts encoded in neurons, and group the ones sharing similar roles.
Since there are no pre-defined concepts, here we assume that annotated labels in datasets can be the guidance. 
To further refine such class-level concepts, we take advantage of domain information.
That is, under DG settings where domains broadly vary in styles, some domain-specific concepts often emerge in each class\footnote{In PACS dataset, for example, the \textit{dog} images under domain \textit{Phote} are often delineated with the fine-grained concept \textit{fur}; whereas, in other domains such as \textit{skeleton} and \textit{cartoon}, such a concept rarely shows.}.
In this sense, our goal is to \textit{refine class-level coarse concepts under different domains} and then \textit{group neurons pertaining to such class-domain conditioned concepts} to get the neuron clusters~~$\mathcal{U}^*$.



We start by constructing a set of class-domain conditioned clusters $U_{c,s}=\{u^{c,s}_i\}_{i=1}^K$, where each cluster $u_i^{c,s}$ comprises neurons identifying a domain-specific concept in class $c$ under domain $s$. 
Inspired by the recent findings that neurons detecting the same concepts often co-activate many common images~\cite{NACT:ConceptEvo,NACT:NeuroCartography}, we build $U_{c,s}$ via \textit{grouping neurons which share a set of top-activated images}. 
In detail, given a set of training samples $X_{c,s}$ under domain $s$ and class $c$, three steps are performed:

\begin{itemize}
	\item \textit{Step1 (Collecting)}: By treating feature elements as neuron activation states, we first collect a neuron activation matrix $\mathbf{Z}_{c,s} = [f_\theta(x_1^{c,s}), ..., f_\theta(x_{|X_{c,s}|}^{c,s})]\in \mathbb{R}^{N\times |X_{c,s}|}$ through the feature extractor $f_\theta$, where $N$ indicates the number of neurons in the last feature layer.
	
	\item \textit{Step2 (Filtering)}: Following the prevalent formulation of neuron activation ~\cite{NACT_System:PathCoverage,NACT_DG:NeuronCoverage,NACT:PNAS_Dissect,NACT:MILAN,NACT:PAMI_Dissect}, we introduce a threshold $t$ to flag activated neurons in $\mathbf{Z}_{c,s}$, whereby neurons with outputs larger than $t$ would be marked as 1 (activated); otherwise, 0 (in-activated).
	\item \textit{Step3 (Clustering)}: 
	After the filtering process,  $n$-th row in $\mathbf{Z}_{c,s}$ would correspond to the set of images that activate $n$-th neuron (which we would refer to as the \textit{stimuli set} of $n$-th neuron).
	In this sense, we finally apply unsupervised clustering (\eg, $K$-Means) to group neurons based on their stimuli sets, forming preliminary clusters $U_{c,s}$.
\end{itemize}

\begin{algorithm}[t]
	\centering
	\caption{Neuron Summarization}
	\label{alg:neuron_summarization}
		\begin{algorithmic}[1] %
			\fontsize{9}{10.5}\selectfont  

			\REQUIRE ~~~~$f_\theta$ $\to$ feature extractor;~~$F_{\theta,\phi}$ $\to$ DNN model;\\
			\quad~~~~~~{$t$} $\to$ threshold for neuron activation; \\
			\quad~~~~~~$\mathcal{D}=\{D_s\}_{s=1}^{S}$ $\to$ training dataset with $C$ classes and $S$ domains.\\
			
			\ENSURE ~$\mathcal{U}^*$ $\to$ concept clusters where each cluster consists of neurons identifying the same visual concepts.\\
			\STATE $\triangleright$ \textit{\CommetBlue{Main procedure starts}}
			\STATE $\mathcal{U} = \{\}$
			\FOR{$c \leftarrow 1, ..., C$}
			\STATE ${U}_c = \{\}$
			\FOR{$s \leftarrow 1, ..., S$}
			\STATE $\triangleright$ \textit{\CommetBlue{Step1: Collecting neuron~states}}\\
			\STATE $X_{c,s}: = \{x_i^{c,s} \in D_s| y_i^{c,s} = c \land F_{\theta, \phi}(x_i^{c,s}) = c\}$
			\STATE $\mathbf{Z}_{c,s} = [f_\theta(x_1^{c,s}), ..., f_\theta(x_{|X_{c,s}|}^{c,s})] $ 
			\STATE $\triangleright$ \textit{\CommetBlue{Step2: Filtering activated neurons}} 
			\STATE \actcond ~$= (\mathbf{Z}_{c,s}> t)$
			\STATE $\mathbf{Z}_{c,s}[$\actcond$] = 1$;~~ $\mathbf{Z}_{c,s}[\lnot$\actcond$] = 0$
			\STATE $\triangleright$ \textit{\CommetBlue{Step3: Clustering neurons unsupervisedly}}\\
			\STATE ${U}_{c,s} = \mathrm{K}$-$\mathrm{Means}(\mathbf{Z} _{c,s})$ 
			\STATE ${U}_c = {U}_c \cup {U}_{c,s}$
			\ENDFOR
			\STATE Compute $U_c^*$ by merging domain-level overlapped clusters in $U_c$ based on Equation~(\ref{equation:jaccard})
			\STATE $\mathcal{U} = \mathcal{U} \cup U_c^*$
			\ENDFOR\\
			\STATE Compute $\mathcal{U}^*$ by merging class-level overlapped clusters in $\mathcal{U}$ based on Equation~(\ref{equation:jaccard})
			\STATE \textbf{return}~~$\mathcal{U}^*$
		\end{algorithmic}
\end{algorithm}

%



Since different domains may enjoy similar visual concepts, we then merge overlapped clusters across domains. In particular, following~\cite{NACT:NeuroCartography}, for each class $c$, we calculate Jaccard similarity on the union set of clusters $U_{c} = \cup_{s \in S}U_{c,s}$,
\begin{equation}
	\label{equation:jaccard}
	{Jaccard}(u_{i}^{c},u_{j}^{c})= \frac{|u_{i}^{c} \cap u_{j}^{c}|}{|u_{i}^{c} \cup u_{j}^{c}|},
\end{equation}
where $u_{i}^{c} \in U_{c}$ and $u_{j}^{c} \in U_{c}$ denote clusters under class $c$. The clusters with a similarity score closer to 1 are merged.
After that, we get the refined class-level clusters $U_{c}^*$. 

Likewise, the overlap among class-level concepts is also considered. As such, by performing Equation~\ref{equation:jaccard} again on the $\mathcal{U} = \cup_{c \in C}U_{c}^*$, we get the final set of neuron clusters $\mathcal{U}^*$. 
Lastly, as neurons may not limit to the single-concept detection, we calculate the contribution of neurons in each concept cluster $u_i \in \mathcal{U}^*$. 
In detail, for each neuron $n \in u_i$, we calculate its weight as the overlap in activated~images:
\begin{equation}
	\label{equation:weight_calculation}
	w_n=\frac{|X_n|}{|\cup_{n \in u_i}X_n|},
\end{equation}
where $X_n = \{x | \forall x\in X,f_{\theta}(x)_n > t\}$ is a set of images that activate neuron $n$. $X$ denotes all training samples. 
Note that since CoCo works in a finetuning fashion when neuron concepts evolve slowly~\cite{NACT:PAMI_Dissect}, we typically update $\mathcal{U}^*$ every 1,000 steps, with a maximum of 5,000 training steps.

{
	\bfstart{Concept Contrast}
	After modeling the relations among neurons, we then perform concept-level contrastive learning by abstracting concept activation vectors (CAVs) from feature embeddings.
	In particular, we define $i$-th element of CAV $\boldsymbol{v}$ as ${v}_i$, which connects with the activation state of $i$-th concept identified in $\mathcal{U}^*$.
	In this sense, given a feature vector $\boldsymbol{z}$, we calculate ${v}_i$ as the weighted sum of all neuron states in $i$-th concept cluster $u_i \in \mathcal{U}^*$, 
	\begin{equation}
		v_i = \sum_{n \in u_i}{w_n z_n},
	\end{equation}
	where $n$ is a neuron in concept cluster $u_i$, and $w_n$ denotes its corresponding weight. $z_n$ is the $n$-th element in the feature vector, which is also treated as activation state of neuron $n$.
	As the philosophy behind the CoCo is to contrast concept states, we simply replace all feature vectors $\boldsymbol{z}$ by $\boldsymbol{v}$ in Eq.~(\ref{equation:feat_contrast}), and then the concept-level contrastive loss $\mathcal{L}_c$ can be obtained.



	

	
	\bfstart{Discussions} 
	It is worth noting that due to domain gaps in DG settings, a large variance over same-class samples often challenges the learning of class-level invariant features~\cite{CL&DG:PCL}. 
	The direct consequence can be the loss of diversity in feature representations, which could lead to the overfitting problem in generalization~\cite{CL&DG:CSG}. 
	However, tackling this issue is always non-trivial. 
	Our approach CoCo can be considered as a relaxation on the contrastive feature learning in DG, which overcomes this problem from a new view: learning \textit{invariant concepts} instead of \textit{invariant features}. 
	Since concept comparison relaxes the constraints on feature elements, the diversity of learned features can be improved.
	Lastly, as there are no extra parameters imposed by CoCo, it allows easy integration into any contrastive method in DG, enabling enhanced feature learning.
}
\section{Experiments}
\label{sec:experiments}
In this section, we first detail the implementation of our CoCo (Section~\ref{sec:implementation}), and then
introduce the feature diversity metric, which is utilized to evaluate learned invariant features (Section~\ref{sec:exp_div_metric}). In Section~\ref{sec:exp_domainbed}, we demonstrate the effectiveness of our CoCo on the four datasets of DomainBed benchmark~\cite{Setup:DomainBed}. The further experiments on the task of synthetic-to-real generalization~\cite{VisDA_Baseline:ASG} are also elaborated, as this task relies on the diversity of learned features (Section~\ref{sec:exp_synthetic_to_real}).


\subsection{Implementation Details}
\label{sec:implementation}
\bfstart{Training} 
Following prevalent formulations of DG approaches~\cite{Baseline:IRM,CL&DG:CSG,CL&DG:Unified}, we train models comprising of a ImageNet-pretrained feature extractor (\eg, ResNet50~\cite{tech:ResNet}), and a predictor (\eg, MLP). 
The whole training process consists of two stages: 1) we first initialize a model trained with the classical ERM strategy~\cite{Baseline:ERM}; and then 2) we fine-tune the model with CoCo-based contrastive loss $\mathcal{L}_c$, where the clustering is typically performed every 1000 training steps.
Most of our implementations are built upon the DomainBed official code using PyTorch\footnote{https://github.com/facebookresearch/DomainBed.}. 
For Synthetic-to-Real generalization, we base our implementation on the source code of CSG\footnote{https://github.com/NVlabs/CSG.}.


\bfstart{Hyperparameters} 
For all employed contrastive methods, we utilize the default hyperparameters of the original paper.
As aforementioned, the implementation of CoCo involves two hyperparameters: the threshold $t$ and the number of clusters $K$. Following~\cite{NACT:PNAS_Dissect}, we dynamically set $t$ as the top 1\% quantile level of neuron outputs. The search space for $K$ is [1, 2, 5, 10].
Besides, we discard neurons with activation times below $\lambda |X_{c,s}|$ to account for abnormal neuron activations. $\lambda$ is the minimum proportion of activated samples required for a valid neuron, and the corresponding search space is [1\%, 5\%, 10\%, 50\%].
All hyperparameters are randomly searched following the protocol of DomainBed~\cite{Setup:DomainBed}, thereby facilitating method adaptation.
Following~\cite{NACT:NeuroCartography}, we merge clusters with a Jaccard similarity greater than 0.8 to address overlapping clusters.

\begin{figure}
	[t]
	\centering 
	\includegraphics[width=0.99\columnwidth]{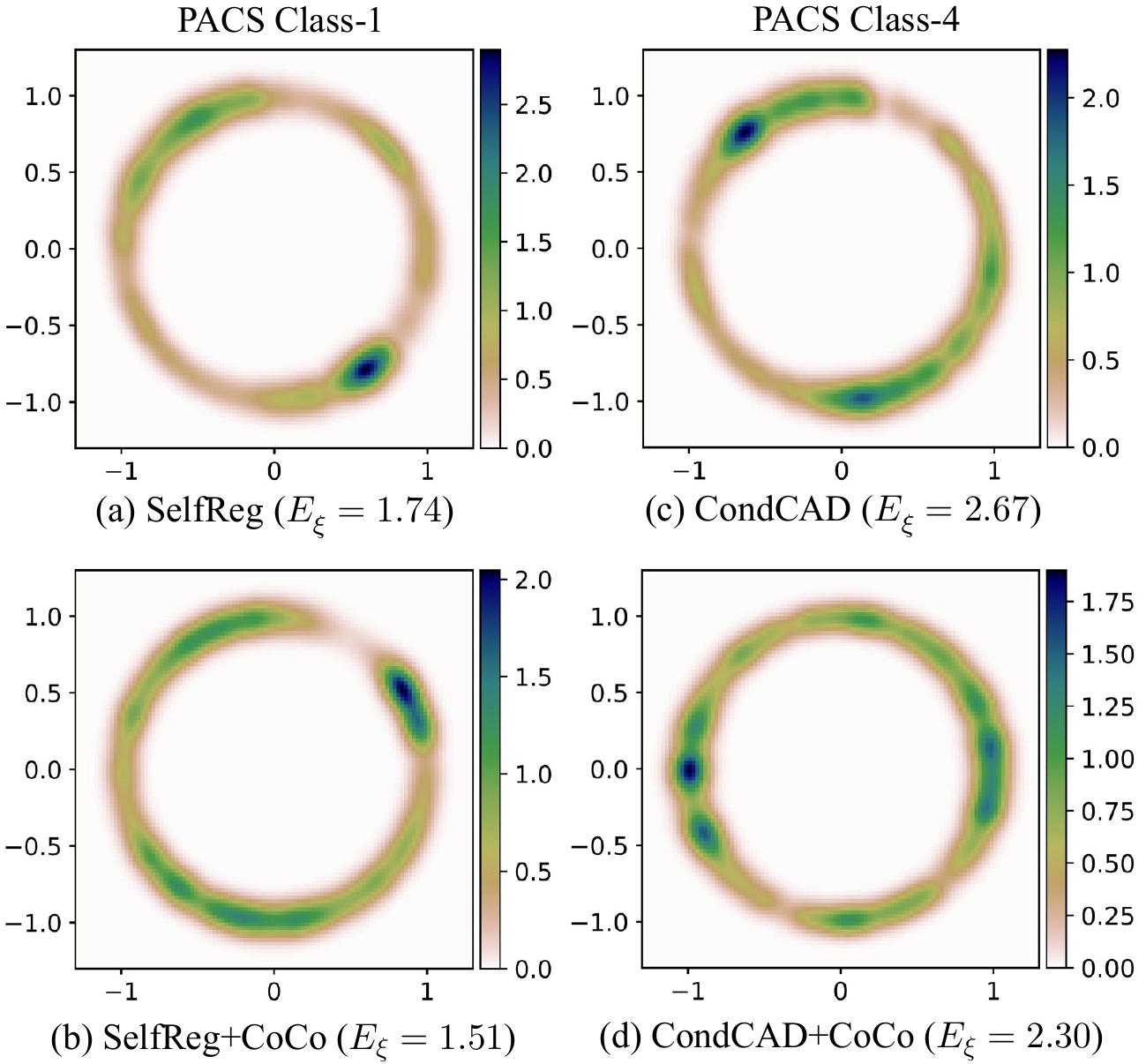} %
	\caption{Comparison of class-level feature diversity between feature contrast methods and our CoCo. We visualize the class-level features on the PACS dataset with Gaussian kernel density estimation (KDE) on 2-dim sphere, where darker areas denote more dense features. $E_{\xi}$ represents the hyperspherical energy of features, which would be lower if the features are more diverse.}
	\label{fig:kde_feat_div}
\end{figure}

\subsection{Feature Diversity Measurement}
\label{sec:exp_div_metric}
While contrastive methods are effective to learn invariant features in DG, they have long been argued with the problem of representation collapse~\cite{Analysis:DimCollapse,CL&Base:BYOL}, thus limiting the diversity of learned features. 
Following~\cite{CL&DG:CSG}, we employ the hyperspherical potential energy~\cite{tech:diversity_energe} to quantitatively measure the diversity of learned invariant features, 
\begin{equation}
		E_{\xi}  = \sum_{i}^{N} \sum_{j \ne i}^{N} e_{{\xi}}(\boldsymbol{z}_i, \boldsymbol{z}_j)=
		\begin{cases} 
			~||\boldsymbol{z}_i - \boldsymbol{z}_j ||^{-s},~s>0 \\ 
			~\mathrm{log}(||\boldsymbol{z}_i - \boldsymbol{z}_j ||^{-1}), s=0,
		\end{cases}
\end{equation}
where $\xi$ denotes the power factor, and $N$ is the number of data samples. A lower potential energy $E_{\xi}$ would represent the more diversified features. We utilize $\xi=0$ in this paper.

\subsection{Results on DomainBed Benchmark}
\label{sec:exp_domainbed}
\bfstart{Settings} We first evaluate our CoCo on the DomainBed benchmark~\cite{Setup:DomainBed}, which is an arguably fairer testbed and rigorously evaluates DG approaches under the same conditions.
As digital images may not contain abundant visual concepts, we experiment with our method on non-MINIST datasets following~\cite{CL&DG:CondCAD}, and four datasets are employed: PACS (4 domains, 9,991 images)~\cite{Dataset:PACS}, OfficeHome (4 domains, 15,588 images)~\cite{Dataset:OfficeHome}, TerraInc (4 domains, 24,788 images)~\cite{Dataset:TerraIncognita}, and DomainNet (6 domains, 586,575 images)~\cite{Dataset:DomainNet}. 

We incorporate CoCo into two SoTA contrastive methods: SelfReg~\cite{CL&DG:SelfReg} and CondCAD~\cite{CL&DG:CondCAD}. 
For a fair comparison, we follow the official source codes and run over all the required processing steps to get results.
To report final scores, we employ training-domain validation sets for best model selection, and average our results over three random seeds.

\begin{table}
	\caption{Test accuracy on PACS dataset~\cite{Dataset:PACS} with ResNet-50 backbone. Results are averaged over 3 random trials.}
	\begin{center}
		\resizebox{0.89\columnwidth}{!}{
			\begin{tabular}{ l | cccc | c }
				\hline &\\[-2ex]
				{Algorithm}   & {Art}           & {Cartoon}           & {Photo}           & {Sketch}           & {Avg.}         \\
				\hline \hline &\\[-2.2ex]
				\hypersetup{colorlinks,linkcolor={red},citecolor={gray},urlcolor={black}}{\color{gray}CORAL~\cite{Baseline:CORAL}} & {\color{gray}88.3} & {\color{gray}80.0} & {\color{gray}97.5} & {\color{gray}78.8} & {\color{gray}86.2} \\
				\hypersetup{colorlinks,linkcolor={red},citecolor={gray},urlcolor={black}}{\color{gray}DANN~\cite{Baseline:DANN}} & {\color{gray}86.4} & {\color{gray}77.4} & {\color{gray}97.3} & {\color{gray}73.5} & {\color{gray}83.6} \\
				MMD~\cite{Baseline:MMD}  			 & 86.1      & 79.4      & 96.6      & 76.5      & 84.6                 \\
				CDANN~\cite{Baseline:CDANN}  		 & 84.6      & 75.5      & 96.8      & 73.5      & 82.6                 \\
				MLDG~\cite{Baseline:MLDG}   		 & 85.5      & 80.1      & 97.4      & 76.6      & 84.9                 \\
				IRM~\cite{Baseline:IRM}  			 & 84.8      & 76.4      & 96.7      & 76.1      & 83.5                 \\
				RSC~\cite{Baseline:RSC}  			 & 85.4      & 79.7      & \textbf{97.6}      & 78.2      & 85.2                 \\
				GroupDRO~\cite{Baseline:GroupDRO}    & 83.5      & 79.1      & 96.7      & 78.3      & 84.4                 \\
				ARM~\cite{Baseline:ARM}  			 & 86.8      & 76.8      & 97.4      & 79.3      & 85.1                 \\
				ERM~\cite{Baseline:ERM}       	 & 84.7      & {80.8}      & {97.2}      & 79.3      & 85.5                 \\
				MTL~\cite{Baseline:MTL}  			 & {87.5}      & 77.1      & 96.4      & 77.3      & 84.6                 \\
				VREx~\cite{Baseline:V-REx}  		 & 86.0      & 79.1      & 96.9      & 77.7      & 84.9                 \\
				And-mask~\cite{Baseline:AND-Mask} 	& 85.3& 79.2& 96.9& 76.2& 84.4\\
				SAND-mask~\cite{Baseline:SAND-Mask} &85.8& 79.2& 96.3& 76.9& 84.6\\
				SagNet~\cite{Baseline:SagNet}  		 & 87.4      & {80.7}      & {97.1}      & {80.0}      & {86.3}                 \\
				Fishr~\cite{Baseline:Fishr} 		& \textbf{88.4}&78.7&97.0&77.8& 85.5\\
				DGRI~\cite{Baseline:DGRI} & / & / & / & / & 84.7 \\
				PCL~\cite{CL&DG:PCL} 	& {87.6}         &79.9         &96.4         &77.4& 85.3 \\
				\hline \hline &\\[-2.2ex]
				CondCAD~\cite{CL&DG:CondCAD} & \textbf{88.4} & 78.4 & 96.8 & 76.4 & 85.0 \\
				\rowcolor{LightGray}
				\quad+CoCo (Ours)  &  {87.9} & {80.7} &  96.9 &  {79.9} &  {86.4}\\
				SelfReg~\cite{CL&DG:SelfReg} & {87.9} & 79.4 & 96.8 & 78.3 & {85.6} \\
				\rowcolor{LightGray}
				\quad+CoCo (Ours)  &  88.1 &  \textbf{81.1} &  96.5 &  \textbf{81.7} &  \textbf{86.9}\\
				\hline
		\end{tabular}}
	\end{center}
	\label{tab:PACS}
\end{table}

\begin{table}
	\caption{Test accuracy on OfficeHome dataset~\cite{Dataset:OfficeHome} with ResNet-50 backbone. Results are averaged over 3 random trials.}
	\begin{center}
		\resizebox{0.89\columnwidth}{!}{
			\begin{tabular}{ l | cccc | c }
				\hline &\\[-2ex]
				{Algorithm}   & {Art}           & {Clipart}           & {Product}           & {Real}           & {Avg.}         \\
				\hline \hline &\\[-2.2ex]
				\hypersetup{colorlinks,linkcolor={red},citecolor={gray},urlcolor={black}}{\color{gray}CORAL~\cite{Baseline:CORAL}} & {{\color{gray}65.3}} & {\color{gray}54.4} & {\color{gray}76.5} & {\color{gray}78.4} & {\color{gray}68.7} \\
				\hypersetup{colorlinks,linkcolor={red},citecolor={gray},urlcolor={black}}{\color{gray}DANN~\cite{Baseline:DANN}} & {\color{gray}59.9} & {\color{gray}53.0} & {\color{gray}73.6} & {\color{gray}76.9} & {\color{gray}65.9} \\
				MMD~\cite{Baseline:MMD}  			 & 60.4      & 53.3      & 74.3      & 77.4      & 66.3                 \\
				CDANN~\cite{Baseline:CDANN}  		 & 61.5      & 50.4      & 74.4      & 76.6      & 65.8                 \\
				MLDG~\cite{Baseline:MLDG}   		 & 61.5      & 53.2      & 75.0      & 77.5      & 66.8                 \\
				IRM~\cite{Baseline:IRM}  			 & 58.9      & 52.2      & 72.1      & 74.0      & 64.3                 \\
				RSC~\cite{Baseline:RSC}  			 & 60.7      & 51.4      & 74.8      & 75.1      & 65.5                 \\
				GroupDRO~\cite{Baseline:GroupDRO}    & 60.4      & 52.7      & 75.0      & 76.0      & 66.0                 \\
				ARM~\cite{Baseline:ARM}  			 & 58.9      & 51.0      & 74.1      & 75.2      & 64.8                 \\
				ERM~\cite{Baseline:ERM}       	 & 61.3      & 52.4      & 75.8      & 76.6      & 66.5                 \\
				MTL~\cite{Baseline:MTL}  			 & 61.5      & 52.4      & 74.9      & 76.8      & 66.4                 \\
				VREx~\cite{Baseline:V-REx}  		 & 60.7      & 53.0      & 75.3      & 76.6      & 66.4                 \\
				And-mask~\cite{Baseline:AND-Mask} 		& 59.5& 51.7& 73.9& 77.1& 65.6\\
				SAND-mask~\cite{Baseline:SAND-Mask} & 60.3& 53.3& 73.5& 76.2& 65.8\\
				SagNet~\cite{Baseline:SagNet}  		 & 63.4      & {54.8}      & 75.8      & 78.3      & {68.1}                 \\
				Fishr~\cite{Baseline:Fishr} 		& 62.4& 54.4&76.2& {78.3}&67.8\\
				DGRI~\cite{Baseline:DGRI} & / & / & / & / & {68.6} \\
				PCL~\cite{CL&DG:PCL} 		&{63.9}         &{55.0}         &{76.9}         &78.2         &{68.5}\\
				\hline \hline &\\[-2.2ex]
				CondCAD~\cite{CL&DG:CondCAD} & 61.8 & 53.7 & 76.5 & 77.7 & 67.4 \\
				\rowcolor{LightGray}
				\quad+CoCo (Ours)  &  {63.2} &  54.2 &  76.8 &  78.3 &  {68.1}\\
				SelfReg~\cite{CL&DG:SelfReg} & {63.6} & 53.1 & 76.9 & 78.1 & 67.9 \\
				\rowcolor{LightGray}
				\quad+CoCo (Ours)  &  \textbf{64.7} &  \textbf{55.3} &  \textbf{77.0} &  \textbf{78.6} &  \textbf{68.9} \\
				\hline
		\end{tabular}}
	\end{center}
	\label{tab:Office}
\end{table}

\begin{table}
	\caption{Test accuracy on TerraInc dataset~\cite{Dataset:TerraIncognita} with ResNet-50 backbone. Results are averaged over 3 random trials.}
	\begin{center}
		\resizebox{0.89\columnwidth}{!}{
			\begin{tabular}{ l | cccc | c }
				\hline &\\[-2ex]
				{Algorithm}   & {Loc100}        & {Loc38}         & {Loc43}         & {Loc46}         & {Avg.}         \\
				\hline \hline &\\[-2.2ex]
				\hypersetup{colorlinks,linkcolor={red},citecolor={gray},urlcolor={black}}{\color{gray}CORAL~\cite{Baseline:CORAL}} & {\color{gray}51.6} & {\color{gray}42.2} & {\color{gray}57.0} & {\color{gray}39.8} & {\color{gray}47.6} \\
				\hypersetup{colorlinks,linkcolor={red},citecolor={gray},urlcolor={black}}{\color{gray}DANN~\cite{Baseline:DANN}} & {\color{gray}51.1} & {\color{gray}40.6} & {\color{gray}57.4} & {\color{gray}37.7} & {\color{gray}46.7} \\
				MMD~\cite{Baseline:MMD}  			 & 41.9      & 34.8      & 57.0      & 35.2      & 42.2                 \\
				CDANN~\cite{Baseline:CDANN}  		 & 47.0      & 41.3      & 54.9      & 39.8      & 45.8                 \\
				MLDG~\cite{Baseline:MLDG}   		 & {54.2}      & {44.3}      & 55.6      & 36.9      & {47.7}                 \\
				IRM~\cite{Baseline:IRM}  			 & \textbf{54.6}      & 39.8      & 56.2      & 39.6      & 47.6                 \\
				RSC~\cite{Baseline:RSC}  			 & 50.2      & 39.2      & 56.3      & {40.8}      & 46.6                 \\
				GroupDRO~\cite{Baseline:GroupDRO}    & 41.2      & 38.6      & 56.7      & 36.4      & 43.2                 \\
				ARM~\cite{Baseline:ARM}  			 & 49.3      & 38.3      & 55.8      & 38.7      & 45.5                 \\
				ERM~\cite{Baseline:ERM}       	 & 49.8      & 42.1      & 56.9      & 35.7      & 46.1                 \\
				MTL~\cite{Baseline:MTL}  			 & 49.3      & 39.6      & 55.6      & 37.8      & 45.6                 \\
				VREx~\cite{Baseline:V-REx}  		 & 48.2      & 41.7      & 56.8      & 38.7      & 46.4                 \\
				And-mask~\cite{Baseline:AND-Mask} 	&50.0&40.2&53.3&34.8&44.6\\
				SAND-mask~\cite{Baseline:SAND-Mask} 	&45.7&31.6&55.1&39.0&42.9\\
				SagNet~\cite{Baseline:SagNet}  		 & 53.0      & 43.0      & \textbf{57.9}      & {40.4}      & {48.6}                 \\
				Fishr~\cite{Baseline:Fishr} 		&50.2&{43.9}&55.7&39.8&47.4\\
				DGRI~\cite{Baseline:DGRI} & / & / & / & / & 47.8 \\
				PCL~\cite{CL&DG:PCL} 	&{53.5}         &41.2         &55.3         &33.7         &45.9\\
				\hline \hline &\\[-2.2ex]
				CondCAD~\cite{CL&DG:CondCAD} & {54.1} & 42.8 & 56.3 & 36.4 & 47.4 \\
				\rowcolor{LightGray}
				\quad+CoCo (Ours)  &  53.9 &  42.0 &  {57.2} &  \textbf{41.4} &  {48.6} \\
				SelfReg~\cite{CL&DG:SelfReg} & 48.8 & 41.3 & {57.3} & 40.6 & 47.0 \\
				\rowcolor{LightGray}
				\quad+CoCo (Ours)  &  {54.3} &  \textbf{44.8} &  \textbf{57.9} &  {40.8} &  \textbf{49.4}\\
				\hline
		\end{tabular}}
	\end{center}
	\label{tab:Terra}
\end{table}

\begin{table}
	\begin{center}
		\caption{Test accuracy on DomainNet dataset~\cite{Dataset:DomainNet} with ResNet-50 backbone. Results are averaged over 3 random trials.}
		\resizebox{0.99\columnwidth}{!}{
			\begin{tabular}{l| cccccc |c}
				\hline &\\[-2ex]
				{Algorithm}   & {Clip}        & {Info}        & {Paint}       & {Quick}       & {Real}        & {Sketch}      & {Avg}         \\
				\hline \hline &\\[-2.2ex]
				\hypersetup{colorlinks,linkcolor={red},citecolor={gray},urlcolor={black}}{\color{gray}CORAL~\cite{Baseline:CORAL}} & {\color{gray}59.2} & {\color{gray}19.7} & {\color{gray}46.6} & {\color{gray}13.4} & {\color{gray}59.8} & {\color{gray}50.1} & {\color{gray}41.5} \\
				\hypersetup{colorlinks,linkcolor={red},citecolor={gray},urlcolor={black}}{\color{gray}DANN~\cite{Baseline:DANN}} & {\color{gray}53.1} & {\color{gray}18.3} & {\color{gray}44.2} & {\color{gray}11.8} & {\color{gray}55.5} & {\color{gray}46.8} & {\color{gray}38.3}\\
				MMD~\cite{Baseline:MMD}  			 & 32.1      & 11.0      & 26.8      & 8.7       & 32.7      & 28.9      & 23.4                 \\
				CDANN~\cite{Baseline:CDANN}  		 & 54.6      & 17.3      & 43.7      & 12.1      & 56.2      & 45.9      & 38.3                 \\
				MLDG~\cite{Baseline:MLDG}   		 & {59.1}      & 19.1      & 45.8      & {13.4}      & {59.6}      & 50.2      & 41.2                 \\
				IRM~\cite{Baseline:IRM}  			 & 48.5      & 15.0      & 38.3      & 10.9      & 48.2      & 42.3      & 33.9                 \\
				RSC~\cite{Baseline:RSC}  			 & 55.0      & 18.3      & 44.4      & 12.2      & 55.7      & 47.8      & 38.9                 \\
				GroupDRO~\cite{Baseline:GroupDRO}    & 47.2      & 17.5      & 33.8      & 9.3       & 51.6      & 40.1      & 33.3                 \\
				ARM~\cite{Baseline:ARM}  			 & 49.7      & 16.3      & 40.9      & 9.4       & 53.4      & 43.5      & 35.5                 \\
				ERM~\cite{Baseline:ERM}       	 & 58.1      & 18.8      & 46.7      & 12.2      & {59.6}      & 49.8      & 40.9                 \\
				MTL~\cite{Baseline:MTL}  			 & 57.9      & 18.5      & 46.0      & 12.5      & 59.5      & 49.2      & 40.6                 \\
				VREx~\cite{Baseline:V-REx}  		 & 47.3      & 16.0      & 35.8      & 10.9      & 49.6      & 42.0      & 33.6                 \\
				And-mask~\cite{Baseline:AND-Mask} 		& 52.3& 16.6&41.6&11.3&55.8&45.4&37.2\\
				SAND-mask~\cite{Baseline:SAND-Mask} 	&43.8&14.8&38.2&9.0&47.0&39.9&32.1\\
				SagNet~\cite{Baseline:SagNet}  		 & 57.7      & 19.0      & 45.3      & 12.7      & 58.1      & 48.8      & 40.3                 \\
				Fishr~\cite{Baseline:Fishr} 			&58.2& {20.2}&{47.7}&12.7&{60.3}&{50.8}&{41.7}\\
				DGRI~\cite{Baseline:DGRI} & / & / & / & / & / & / & {41.9} \\
				PCL~\cite{CL&DG:PCL} 	&56.0         &20.1         &47.0         &12.5         &56.9         &48.9         &40.2\\
				\hline \hline &\\[-2.2ex]
				CondCAD~\cite{CL&DG:CondCAD}  & 58.7 & 19.9 & 46.9 & 13.1 & 59.4 & 48.0 & 41.0 \\
				\rowcolor{LightGray}
				\quad+CoCo (Ours) &  {58.8} &  {20.4} &  {48.2} &  {13.5} &  {60.6} &  50.2 & {41.9} \\
				SelfReg~\cite{CL&DG:SelfReg} & 58.5 & \textbf{20.7} & 47.3 & 13.1 & 58.2 & {51.1} & 41.5 \\
				\rowcolor{LightGray}
				\quad+CoCo (Ours)  &  \textbf{59.7} &  {20.4} &  \textbf{49.2} &  \textbf{13.9} &  \textbf{61.1} &  \textbf{51.2}& \textbf{42.6} \\
				\hline
		\end{tabular}}
		\label{tab:DomainNet}
	\end{center}
\end{table}

\bfstart{Effectiveness in Improving Invariant Feature Learning} 
As discussed previously (e.g., Figure~\ref{fig:intro_motivation}), feature contrast learns class-conditional invariant features by imposing constraints on all neurons, which may suffer from the overfitting problem in feature representations. 
In this regard, we first verify such unwanted behavior, and then show the effectiveness of CoCo in improving invariant feature learning.
Specifically, following~\cite{CL&DG:CSG,CL:Align&Uniformity} we plot the class-level feature diversity for two feature contrast methods, SelfReg and CondCAD, with and without CoCo. 
As illustrated in Figure~\ref{fig:kde_feat_div}, the class-level features learned via feature-level contrast methods mostly center on a narrow subspace, which thus causes the collapsed representations with the overfitting problem.
In comparison, by adopting concept-level contrasts (CoCo), feature representations are improved to span the whole space. This further introduces the higher diversity in class-level invariant features (\ie, lower hyperspherical energy $E_{\xi}$), and demonstrates the effectiveness of our CoCo in improving feature learning.

\begin{table}
	\caption{Comparison with SoTA methods on synthetic-to-real generalization. \textit{Hyper-energy} denotes the hyperspherical energy of features, which is lower if features are more diverse.}
	\begin{center}
		\resizebox{0.87\columnwidth}{!}{
			\begin{tabular}{l| c | c}
				\hline &\\[-2ex]
				Algorithm       & Hyper-energy $\downarrow$            & Accuracy $\uparrow$ \\
				\hline \hline &\\[-2ex]
				Oracle on ImageNet & - & 53.3 \\
				Weight $l2$ distance~\cite{VisDA_Baseline:Weight_l2} & 0.401 & 56.4\\
				Synaptic Intelligence~\cite{VisDA_Baseline:Synaptic_Intelligence} & 0.396 & 57.6\\
				Feature $l2$ distance~\cite{VisDA_Baseline:Feat_l2} & 0.334 & 57.1\\
				ASG~\cite{VisDA_Baseline:ASG} & 0.325 & 61.1\\
				\hline \hline &\\[-2ex]
				CSG~\cite{CL&DG:CSG}     & 0.319            & 64.05 \\
				\rowcolor{LightGray}
				\quad+CoCo (Ours)     & \textbf{0.304}            & \textbf{65.86} \\
				\hline
		\end{tabular}}
	\end{center}
	\label{table:visda}
\end{table}

\bfstart{Effectiveness in Improving Generalization Ability} 
We further compare our approaches with recent state-of-the-art (SoTA) methods on four datasets of DomainBed. Table~\ref{tab:PACS}-\ref{tab:DomainNet} illustrate the results, where {\color{gray}gray} color denotes the methods designed for unsupervised domain adaptation, which peek target domains during training.
Through the results, the following observations can be drawn:

Firstly, our SelfReg+CoCo achieves a new state-of-the-art performance across four DomainBed datasets. It even outperforms CORAL~\cite{Baseline:CORAL}, which peeks data from target domains. 

Secondly, CoCo consistently enhances the generalization capability of contrastive methods (SelfReg and CondCAD) across the four DomainBed datasets. 
For instance, by incorporating CoCo into SelfReg and CondCAD, a 2.4\% and 1.2\% improvement can be obtained on the TerraInc dataset, respectively. This highlights the effectiveness of our CoCo.

Thirdly, in comparison with CondCAD, CoCo often introduces a larger improvement over SelfReg. 
We conjecture that the reason lies in the training strategy of SelfReg, which solely utilizes the positive feature alignment during training. This introduces a higher risk of overfitting, as verified by the severe collapse issue in Figure~\ref{fig:kde_feat_div}.
In this sense, incorporating CoCo takes more benefits as overfitting is largely alleviated.




\subsection{Results on Synthetic-to-Real Generalization}
\label{sec:exp_synthetic_to_real}
\bfstart{Settings} 
We also consider the task of synthetic-to-real generalization to further validate the effectiveness of our approach, whereby the training set merely consists of synthetic images rendered by 3D models; whereas the target data is all collected from real images. 
Following~\cite{VisDA_Baseline:ASG}, we conduct experiments on the benchmark: VisDA-17$\rightarrow$COCO. 
In particular, the training and validation sets in VisDA dataset~\cite{Dataset:VisDA} are employed as the source domain and target domain, respectively. 
Similar to the DomainBed experiments, we evaluate our method by incorporating it into a novel feature-contrast approach CSG~\cite{CL&DG:CSG}, and each processing step is carefully followed.






\bfstart{Effectiveness in Improving Generalization Ability} 
Since this task is to adapt synthetically trained models to real images, it has been shown that improving feature diversity can be the key to prevent overfitting and improve model generalization~\cite{CL&DG:CSG,VisDA_Baseline:ASG,VisDA:Road}. 
From this perspective, we measure the feature diversity on target data, and compare our method with recent SoTA in Table~\ref{table:visda}. 
It can be seen that CoCo not only improves diversity of learned invariant features over CSG, but also enhances generalization ability of trained models. This highlights the effectiveness of CoCo again.

\begin{table}[t]
	\caption{SWAD-based results on PACS dataset with ResNet-50.}
	\begin{center}
		\resizebox{0.93\columnwidth}{!}{
			\begin{tabular}{l| cccc |c}
				\hline &\\[-2ex]
				{Algorithm}       & {Art}      & {Cartoon}    & {Photo}    & {Sketch}    & {Avg.}              \\
				\hline \hline &\\[-2ex]
				SWAD~\cite{tech:swad} & 89.3 & 83.4 & {97.3} & 82.5 & 88.1\\
				\quad+PCL~\cite{CL&DG:PCL} & 90.2 & 83.9 & \textbf{98.1} & 82.6 & 88.7\\
				\quad+SupCL~\cite{CL:SupCL}     & {90.1} & {81.8} & 97.6 & {83.2} & {88.2} \\
				\rowcolor{LightGray}
				\quad+SupCL+CoCo & \textbf{91.6} & \textbf{85.1} & 97.6 & \textbf{85.2} & \textbf{89.9} \\
				\hline
		\end{tabular}}
	\end{center}
	\label{table:swad_pacs}
\end{table}

\begin{table}[t]
	\caption{SWAD-based results on TerraInc dataset with ResNet-50.}
	\begin{center}
		\resizebox{0.93\columnwidth}{!}{
			\begin{tabular}{l| cccc |c}
				\hline &\\[-2ex]
				{Algorithm}       & Loc100 & Loc38 & Loc43 & Loc46        & {Avg.}              \\
				\hline \hline &\\[-2ex]
				SWAD~\cite{tech:swad} & 55.4 & 44.9 & {59.7} & 39.9 & 50.0\\
				\quad+PCL~\cite{CL&DG:PCL} & 58.7 & 46.3 & \textbf{60.0} & 43.6 & 52.2\\
				\quad+SupCL~\cite{CL:SupCL}    & {56.6} & {44.6} & 57.1 & {41.1} & {49.8} \\
				\rowcolor{LightGray}
				\quad+SupCL+CoCo     & \textbf{59.9} & \textbf{53.1} & 58.5 & \textbf{44.4} & \textbf{54.0} \\
				\hline
		\end{tabular}}
	\end{center}
	\label{table:swad_terra}
\end{table}

\begin{table*}
	\caption{Architecture analysis on DomainBed benchmark. Two pretrained CLIP models with ResNet50~\cite{tech:ResNet} and ViT-B/32~\cite{tech:ViT} are respectively employed. \textit{Baseline} denotes the linear classifier trained on frozen feature extractor.}
	\begin{center}
		\resizebox{0.78\textwidth}{!}{
			\begin{tabular}{ l l | cccc | c}
				\hline &\\[-2ex]
				{Bakbone}          & {Algorithm} & {PACS}~\cite{Dataset:PACS}             & {OfficeHome}~\cite{Dataset:OfficeHome}       & {TerraInc}~\cite{Dataset:TerraIncognita}   & {DomainNet}~\cite{Dataset:DomainNet}        & {Average}       \\
				\hline \hline &\\[-2ex]
				\multirow{3}[1]{*}{CLIP-ResNet50} 
				& Baseline       & 90.3 $\pm$ 0.2            & 70.6 $\pm$ 0.1            & 29.6 $\pm$ 0.8            & 47.7 $\pm$ 0.0           & 59.6                      \\
				& CondCAD~\cite{CL&DG:CondCAD}      & 90.0 $\pm$ 0.6            & 70.5 $\pm$ 0.3            & 30.3 $\pm$ 0.9            & 45.5 $\pm$ 2.1            & 59.1                      \\
				&\cellcolor{LightGray} \quad+CoCo (Ours)   &\cellcolor{LightGray}90.8 $\pm$ 0.4            &\cellcolor{LightGray}70.8 $\pm$ 0.1            &\cellcolor{LightGray}32.6 $\pm$ 1.2            &\cellcolor{LightGray}47.5 $\pm$ 0.1            &\cellcolor{LightGray}60.4                     \\
				\hline &\\[-2ex]
				\multirow{3}[1]{*}{CLIP-ViT-B/32} 
				& Baseline        & 93.5 $\pm$ 0.8            & 79.4 $\pm$ 0.2            & 37.5 $\pm$ 0.7            & 50.1 $\pm$ 1.1           & 65.1                      \\
				& CondCAD~\cite{CL&DG:CondCAD}            & 93.5 $\pm$ 0.7            & 79.7 $\pm$ 0.2            & 37.4 $\pm$ 1.2            & 51.7 $\pm$ 1.4            & 65.6                      \\
				&\cellcolor{LightGray}\quad+CoCo (Ours)    &\cellcolor{LightGray}\textbf{94.1} $\pm$ 0.3            &\cellcolor{LightGray}\textbf{79.8} $\pm$ 0.1            &\cellcolor{LightGray}\textbf{38.3} $\pm$ 0.5            &\cellcolor{LightGray}\textbf{53.3} $\pm$ 0.1            &\cellcolor{LightGray}\textbf{66.4}                      \\
				\hline
		\end{tabular}}
	\end{center}
	\label{table:archecture_analysis}
	\vspace{-2mm}
\end{table*}

\begin{figure*}
	[t]
	\centering 
	\includegraphics[width=0.85\textwidth]{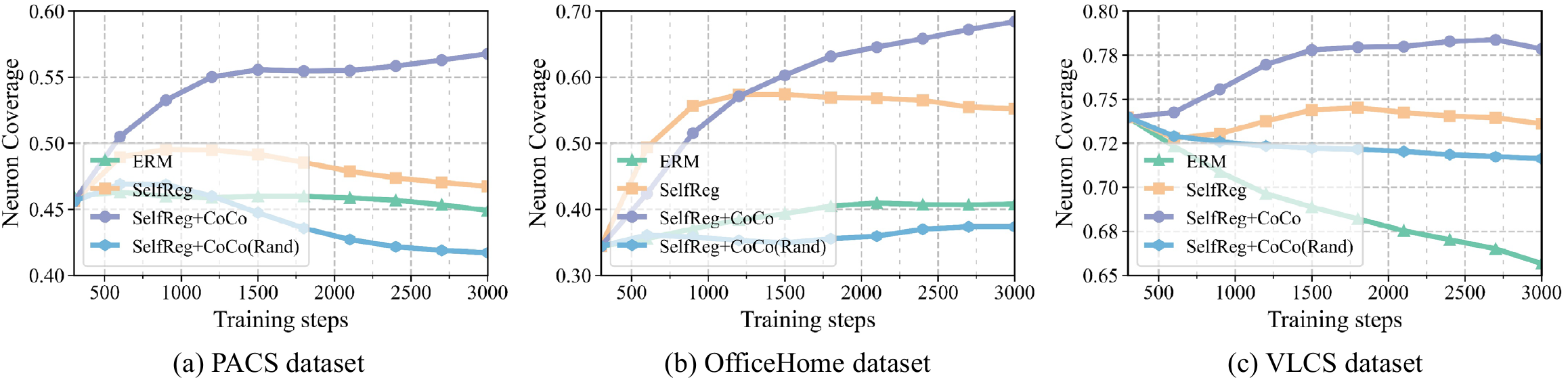} %
	\vspace{-2mm}
	\caption{Neuron coverage analysis on the PACS~\cite{Dataset:PACS}, OfficeHome~\cite{Dataset:OfficeHome}, and VLCS~\cite{Dataset:VLCS} datasets. Following~\cite{NACT_DG:NeuronCoverage,NACT_System:DeepXplore,NACT_System:DeepGauge}, we denote neuron coverage as the proportion of activated neurons under the training set. All curves are obtained by finetuning the same initialized ERM model with 300 pre-training steps. 
		\textit{CoCo (Rand)} implies that neurons in the same concept cluster are randomly selected.}
	\label{fig:exp_coverage}
\end{figure*}

\section{Analysis}
In previous sections, we have shown the effectiveness of our CoCo on two benchmarks. 
In line with these experiments, we provide more analysis about our CoCo in this section.

\subsection{SWAD Analysis} 
Stochastic Weight Averaging Densely (SWAD)~\cite{tech:swad} has recently been widely utilized in various domain generalization approaches~\cite{CL&DG:SelfReg,CL&DG:PCL,Baseline:BatchFormer} as an efficient training technique, which enhances model generalization ability through dense model ensemble. In order to further understand the potential of CoCo, we conduct a series of experiments based on SWAD.
As shown in Table~\ref{table:swad_pacs} and Table~\ref{table:swad_terra}, we implement CoCo on the plain supervised contrastive loss (SupCL)~\cite{CL:SupCL}, and compare it with the recent SoTA, proxy-based contrastive method (PCL)~\cite{CL&DG:PCL}. 
It can be seen that CoCo not only significantly improves SupCL on both PACS and TerraInc datasets, but also outperforms PCL by a large margin. For instance, on the TerraInc dataset, incorporating CoCo results in a +4.2\% improvement in average accuracy for SupCL, surpassing PCL by 1.8\%.
This finding underscores the potential of CoCo to further enhance model generalization ability.

\subsection{Architecture Analysis}
The key to our approach CoCo is dissecting neurons, and grouping the ones identifying the same concepts. 
As this approach hinges on the concept detection ability of neurons, we hypothesize that with a better pretrained model as backbone, CoCo could bring more remarkable improvement. 
To this end, we further experiment with our method on two CLIP models~\cite{tech:Clip} pretrained with ResNet-50~\cite{tech:ResNet} and ViT-B/32~\cite{tech:ViT} over DomainBed. 
As shown in Table~\ref{table:archecture_analysis}, we can see that our CoCo still consistently improves contrastive methods over all datasets, and performs the best on the average accuracy. 
In addition, compared to the results based on ResNet50 in Table~\ref{tab:PACS}-\ref{tab:DomainNet}, it can be found that incorporating CoCo into CLIP-based backbones could cause higher improvement in many cases, \eg, +2.0\% on CLIP-ResNet50 vs.~+0.9\% on ResNet50 over DomainNet.  
Such results confirm our hypothesis and further shows the great potential of CoCo in model generalization.

\begin{figure}
	[t]
	\centering 
	\includegraphics[width=0.9\columnwidth]{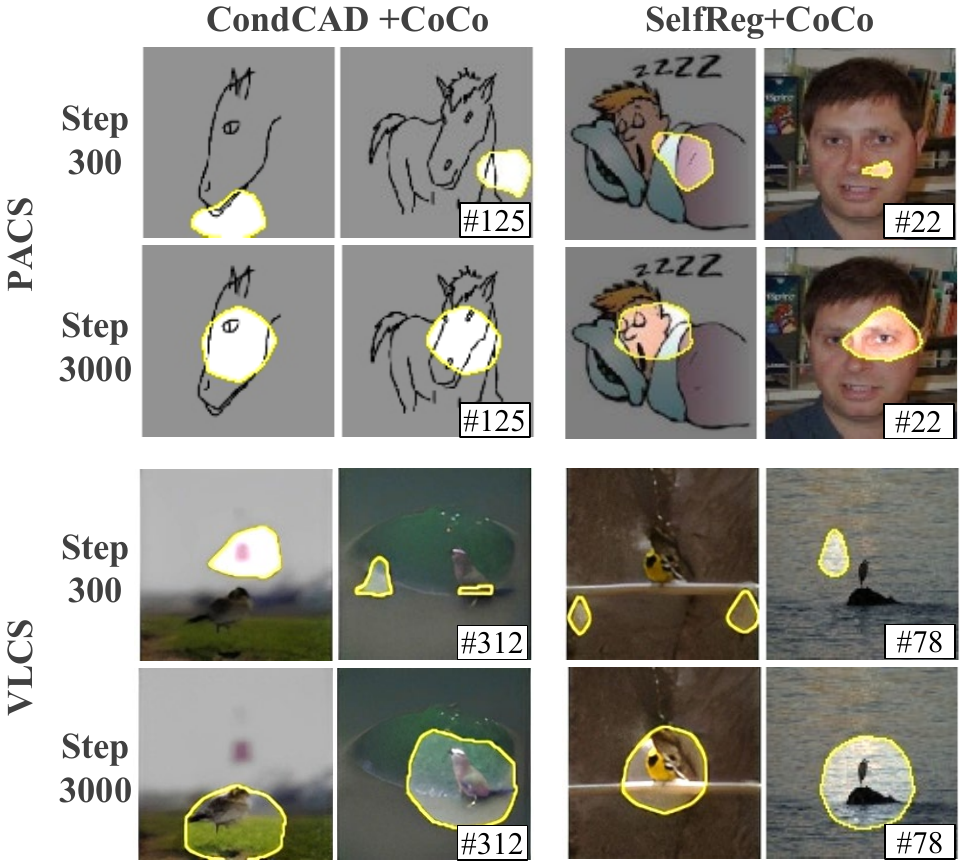} %
	\caption{Concept evolution happens in newly activated neurons. We obtain all neurons from the penultimate layer of the backbone model.}
	\label{fig:coverage_examples}
\end{figure}

\subsection{Neuron Coverage Analysis} 
\bfstart{Neuron Coverage Criteria}
Accuracy as the indicator of model generalization ability has recently been argued with a large variance in the final results~\cite{Setup:DomainBed,CL&DG:CondCAD}. 
In this sense, we take another criterion, neuron coverage~\cite{NACT_System:DeepXplore}, to evaluate the generalization ability of our method.
Following~\cite{NACT_DG:NeuronCoverage,NACT_System:DeepXplore,NACT_System:DeepGauge}, we denote the neuron coverage as the proportion of activated neurons under the training set, 
\begin{equation}
	NCov = \frac{|\{n | \forall x \in X, f_{\theta}(x)_n >t \}|}{N},
\end{equation}
where $X$ denotes all training samples, $n$ is the neuron id, and $N$ is the number of neurons in the last feature layer. The high level idea is that if a model performs with large neuron coverage during training, there would be fewer undetected bugs (e.g., misclassification) triggered by unseen samples. 

\bfstart{Results}
Figure~\ref{fig:exp_coverage} illustrates the neuron coverage curve of our CoCo based on SelfReg. 
For a fair comparison, we implement all baseline methods on the same pretrained ERM model, and then compare the change in the neuron coverage. 
As can be seen, by introducing our CoCo, the coverage ratio on all datasets present an increasing trend, which further achieves the highest one among all the methods. 
This proves the effectiveness of our CoCo again.
In addition, if we ablate the CoCo with random concept clusters, a significant gap in the coverage ratio can be observed, which highlights the advantages of our concept clustering method. 
Lastly, it can also be seen that the results presented by neuron coverage are consistent with the accuracy in Table~\ref{tab:PACS}, \ie, SelfReg+CoCo $>$ SelfReg $>$ ERM.
This indicates that coverage ratio can be a reasonable measurement of model generalization ability.

To further understand the benefits of a broader neuron coverage, we dissect newly activated neurons by comparing their activation map with their inactivated versions. We provide the visualizations in Figure~\ref{fig:coverage_examples}. 
As can be seen, by incorporating CoCo, inactivated neurons could evolve as concept detectors and identify the meaningful information in images, e.g., the unit 312 in CondCAD+CoCo evolves to detect birds in step 3000.
This is in stark contrast to their inactivated period (\eg, step 300), when they merely encode insignificant regions, and thus can easily mislead a model if they are activated during test phases.
As such, we can confirm our previous conclusions that larger coverage ratio introduced by CoCo promotes the model generalization ability.








\subsection{Clustering Method Analysis} 
As illustrated in Algorithm~\ref{alg:neuron_summarization}, our neuron clustering strategy takes advantage of annotated information, such as class and domain information, to model fine-grained concepts. 
To further evaluate the effectiveness of this approach, we run additional ablations.
Table~\ref{table:clustering_analysis} illustrates the results, where we perform an ablation study by comparing the clustering method of CoCo with and without the use of annotations.
As can be observed, CoCo achieves superior results when both domain and class annotations are employed. Additionally, even in the absence of annotations, CoCo still improves the baseline contrastive method on the average accuracy. 
These findings demonstrate the effectiveness of CoCo and its clustering method.


\begin{table}
	\caption{Ablation studies on clustering methods of CoCo over \\PACS dataset~\cite{Dataset:PACS}. Backbone: ResNet-50. }
	\begin{center}
		\resizebox{0.99\columnwidth}{!}{
			\begin{tabular}{l| cccc |c}
				\hline &\\[-2ex]
				{Algorithm}       & {Art}      & {Cartoon}    & {Photo}    & {Sketch}    & {Avg}              \\
				\hline \hline &\\[-2ex]
				SelfReg~\cite{CL&DG:SelfReg} & {87.9} & 79.4 & \textbf{96.8} & 78.3 & {85.6} \\
				\quad+CoCo (w/o class \& domain) & \textbf{90.3} & 77.7 & {96.3} & 78.7 & 85.8\\
				\quad+CoCo (w/o domain) &  {88.3} &  {79.8} &  96.3 &  {80.8} &  {86.3}\\
				\quad+CoCo (w/o class) & 89.7 & 76.7 & \textbf{96.8} & \textbf{\textbf{83.4}} & 86.6\\
				\rowcolor{LightGray}
				\quad+CoCo (Ours) &  88.1 &  \textbf{81.1} &  96.5 &  {81.7} &  \textbf{86.9}\\
				\hline
		\end{tabular}}
	\end{center}
	\label{table:clustering_analysis}
\end{table}

\begin{figure}
	[t]
	\centering 
	\includegraphics[width=0.97\columnwidth]{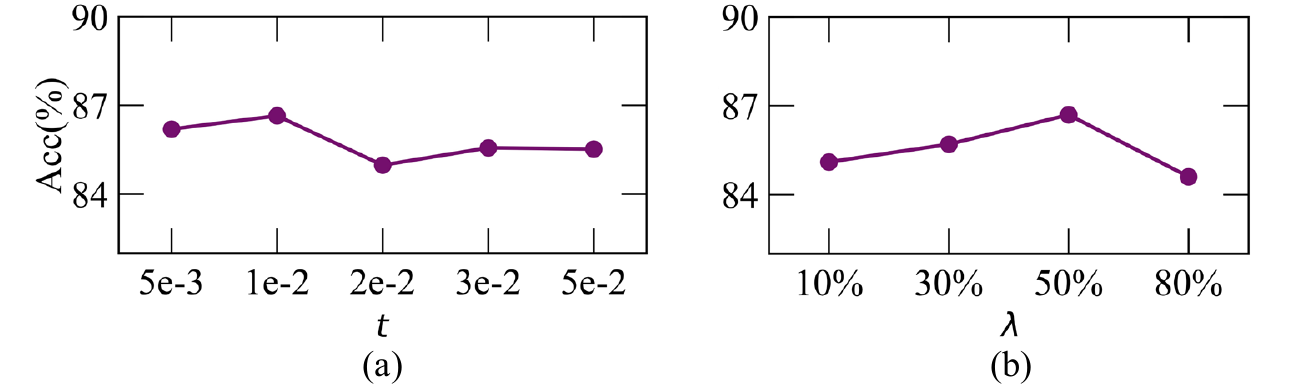} %
	\vspace{-3mm}
	\caption{Ablation studies on the different activation threshold ($t$) and ratio of activated images ($\lambda$) over PACS dataset~\cite{Dataset:PACS}. Backbone: ResNet-50.}
	\label{fig:param_plot}
\end{figure}

\subsection{Hyperparameter Analysis} 
We also run a series of experiments to study the hyperparameters of CoCo, \ie, the activation threshold $t$, and the ratio of activated images $\lambda$ for discarding noisy neurons.
The results are shown in Figure~\ref{fig:param_plot}. 
It can be seen that a relatively lower activation threshold $t$ often makes CoCo perform better. 
This can be attributed to the fact that a larger threshold would overlook more informative images that activate neurons, thus limiting the similarity computation between neurons.
In addition, we can further observe that CoCo achieves best performance when the ratio $\lambda=50\%$. 
We consider the reason is that a larger $\lambda$ would discard many meaningful neurons; whereas, a smaller one would consider more abnormally activated neurons. Both scenarios hinder the effective clustering of neurons, which in turn degrades the performance of CoCo.

\subsection{Computational Cost Analysis} 
As previously discussed, CoCo performs neuron clustering in an infrequent manner, \ie, clustering is executed every 1,000 training steps. 
This infrequent process effectively reduces the extra computational burden associated with CoCo.
To gain a quantitative understanding of such computational cost, we compare the computation time of feature contrast method (CondCAD) and our concept contrast (CondCAD+CoCo). 
Table~\ref{table:computation_cost} illustrates the results on a single RTX 3090.
As can be seen, the extra computation time introduced by CoCo is merely 7.6 seconds for every 1,000 training steps. 
This amounts to approximately 3.6\% additional running time when compared to the feature contrast method.
Given such minimal impact on computational resources, CoCo proves to be an efficient technique for enhancing the model generalization ability. 

\begin{table}
	\caption{Comparison of computation time (1,000 steps) between \\Feature Contrast and Concept Contrast. }
	\begin{center}
		\resizebox{0.82\columnwidth}{!}{
			\begin{tabular}{c | c}
				\hline &\\[-2ex]
				{Operation}       &  Computation Time (sec)     \\
				\hline \hline &\\[-2ex]
				Feature Contrast Computation & 212.6s \\
				
				\hline \hline &\\[-2ex]
				Neuron Summarization & 7.8s \\
				Concept Contrast Computation & 212.4s \\
				Total & 220.2s \\
				\hline
		\end{tabular}}
	\end{center}
	\label{table:computation_cost}
\end{table}


%

\section{Conclusions}
This paper explores the contrastive feature alignment in domain generalization from a neuron activation view. We presented a simple yet effective approach, Concept Contrast (CoCo), to relax element-wise feature alignments by contrasting high-level concepts encoded in neurons. Through the comparison with feature contrast methods, we demonstrated the capability of our CoCo to alleviate overfitting problem in invariant feature learning. 
The results on DomainBed and synthetic-to-real generalization further show that CoCo not only consistently improves model performance, but also contributes to a broader neuron coverage for model generalization. 


{
	\bibliographystyle{IEEEtran}
	\bibliography{mybib}
}

\newpage

%

%
%

\vfill

\end{document}